%% file: root.tex
\documentclass[conference]{IEEEtran}
\usepackage{algorithm}
\usepackage{algpseudocode}
\usepackage{amsmath}
\usepackage{amssymb}
\usepackage[hyperfootnotes=false,colorlinks=true,linkcolor=blue,citecolor=blue,urlcolor=blue]{hyperref}
\usepackage{graphicx}
\usepackage{subcaption}
\usepackage[table]{xcolor} 
\usepackage{array} 
\usepackage{tabularx}
\usepackage{xcolor,colortbl}
\usepackage{geometry}

\makeatletter
\newcommand{\firstpagetopmargin}{4.3mm} 
\makeatother

\begin{document}

\title{
    \vspace{\firstpagetopmargin} 
    \textbf{\fontsize{14}{12}\selectfont Incremental Mapping with Measurement Synchronization \& Compression}
}

\author{
    Mark Griguletskii\textsuperscript{1}, Danil Belov\textsuperscript{1}, Pavel Osinenko\textsuperscript{1}
}

\maketitle

\begingroup
\renewcommand{\thefootnote}{}
\footnotetext{
    \vspace{-1mm} 
    \hspace{-3mm} \begin{minipage}{0.99\linewidth}
        \footnotesize
        \textsuperscript{1}Digital Engineering Department, Skolkovo Institute of Science and Technology, Moscow, Russia. Emails: [\texttt{Mark.Griguletskii, Danil.Belov, P.Osinenko}]\texttt{@skoltech.ru}.
    \end{minipage}
}
\endgroup

\setcounter{footnote}{1}

\input{chapters/abstract}


\input{chapters/introduction}
\input{chapters/related_work}
\input{chapters/methodology}
\input{chapters/experiments}
\input{chapters/discussion}
\input{chapters/conclusion}
\bibliographystyle{./IEEEtran/bibtex/IEEEtran}
\bibliography{references}
\input{chapters/appendix}

\end{document}

%% file: chapters/abstract.tex
\begin{abstract}
	Modern autonomous vehicles and robots utilize versatile sensors for localization and mapping. The fidelity of these maps is paramount, as an accurate environmental representation is a prerequisite for stable and precise localization. Factor graphs provide a powerful approach for sensor fusion, enabling the estimation of the maximum a posteriori solution. However, the discrete nature of graph-based representations, combined with asynchronous sensor measurements, complicates consistent state estimation. The design of an optimal factor graph topology remains an open challenge, especially in multi-sensor systems with asynchronous data. Conventional approaches rely on a rigid graph structure, which becomes inefficient with sensors of disparate rates. Although pre-integration techniques can mitigate this for high-rate sensors, their applicability is limited. To address this problem, this work introduces a novel approach that incrementally constructs connected factor graphs, ensuring the incorporation of all available sensor data by choosing the optimal graph topology based on the external evaluation criteria. The proposed methodology facilitates graph compression, reducing the number of nodes (optimized variables) by \(\sim\)30\% on average while maintaining map quality at a level comparable to conventional approaches.
\end{abstract}

%% file: chapters/introduction.tex
\section{Introduction}
\label{sec:intro}

\begin{figure*}[t]
    \centering

    \begin{minipage}[t]{0.34\textwidth}
        \includegraphics[width=\linewidth, height=0.12\textheight]{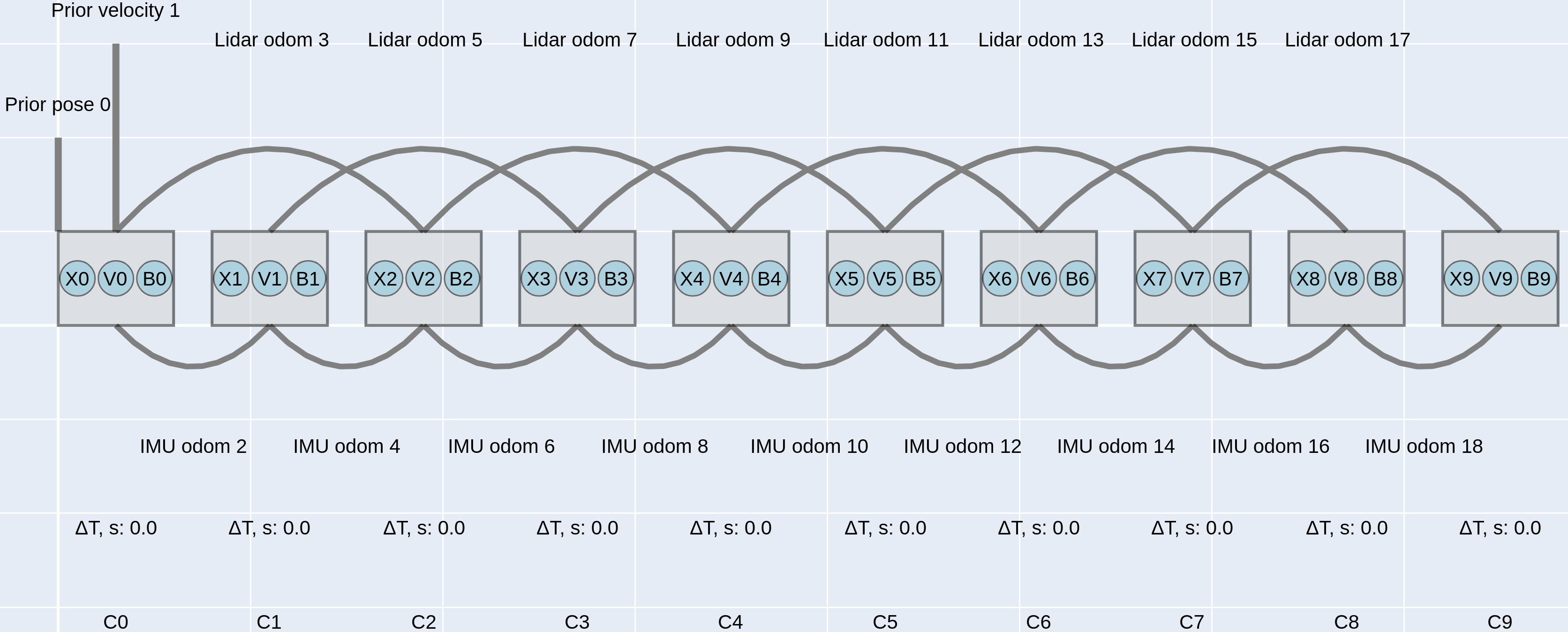}
    \end{minipage}
    \hfill
    \begin{minipage}[t]{0.29\textwidth}
        \includegraphics[width=\linewidth, height=0.15\textheight]{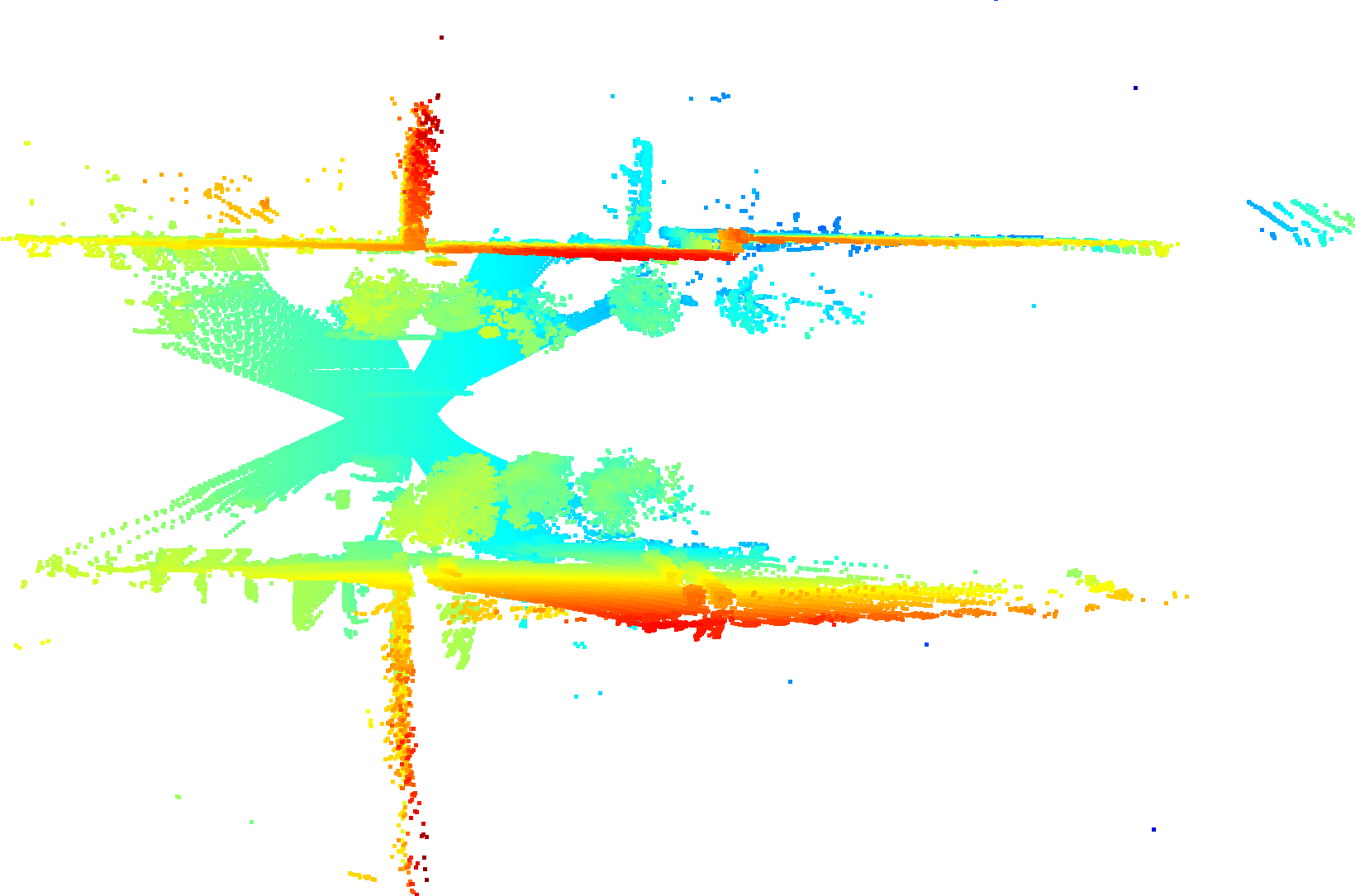}
    \end{minipage}
    \hfill
    \begin{minipage}[t]{0.349\textwidth}
        \includegraphics[width=\linewidth, height=0.12\textheight]{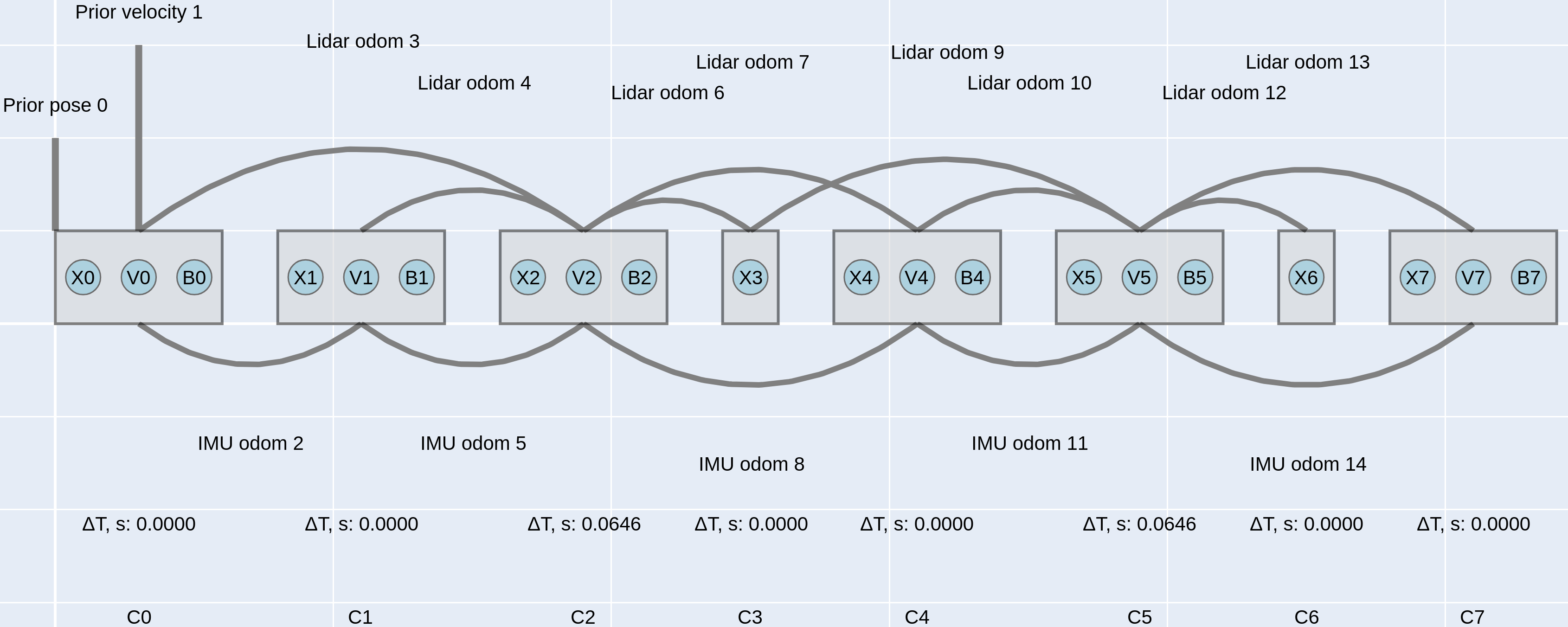}
    \end{minipage}

    \begin{minipage}[t]{0.34\textwidth}
        \includegraphics[width=\linewidth, height=0.12\textheight]{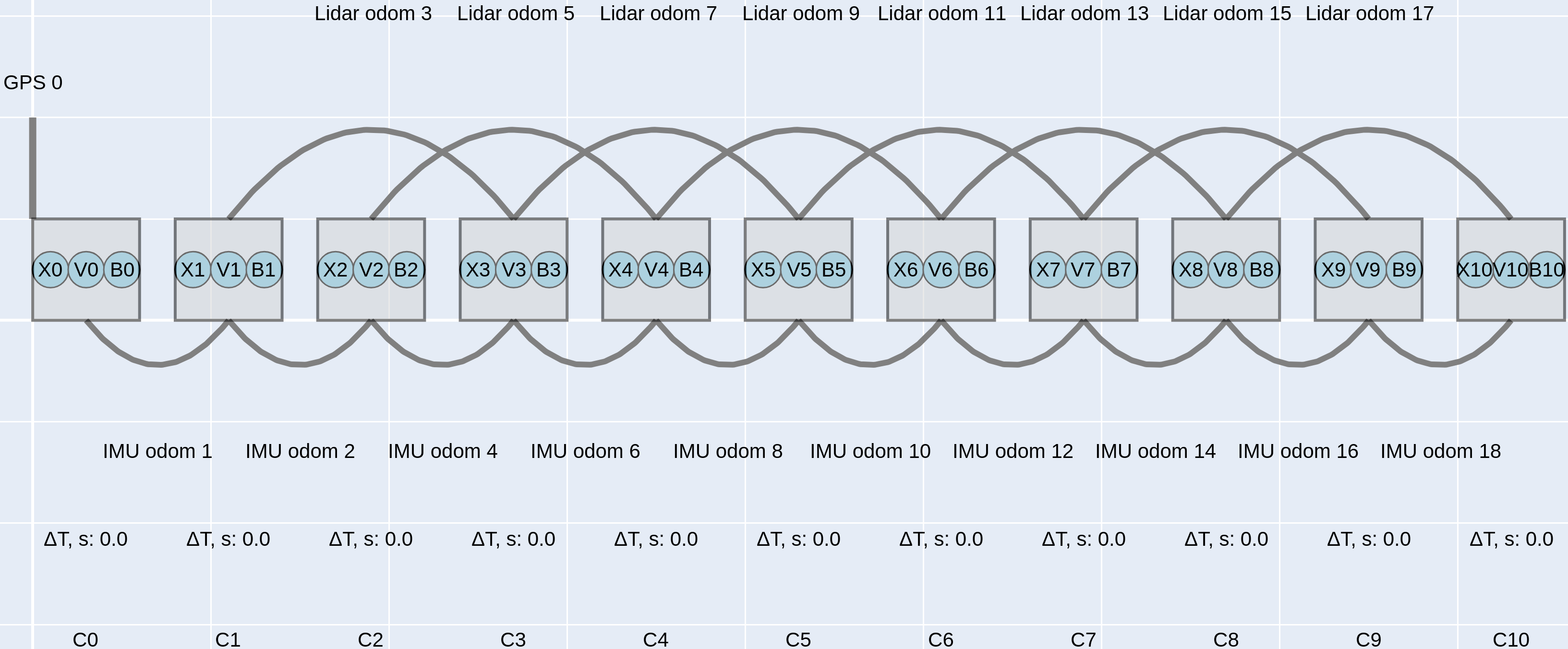}
        \subcaption{Base}
    \end{minipage}
    \hfill
    \begin{minipage}[t]{0.29\textwidth}
        \includegraphics[width=\linewidth, height=0.15\textheight]{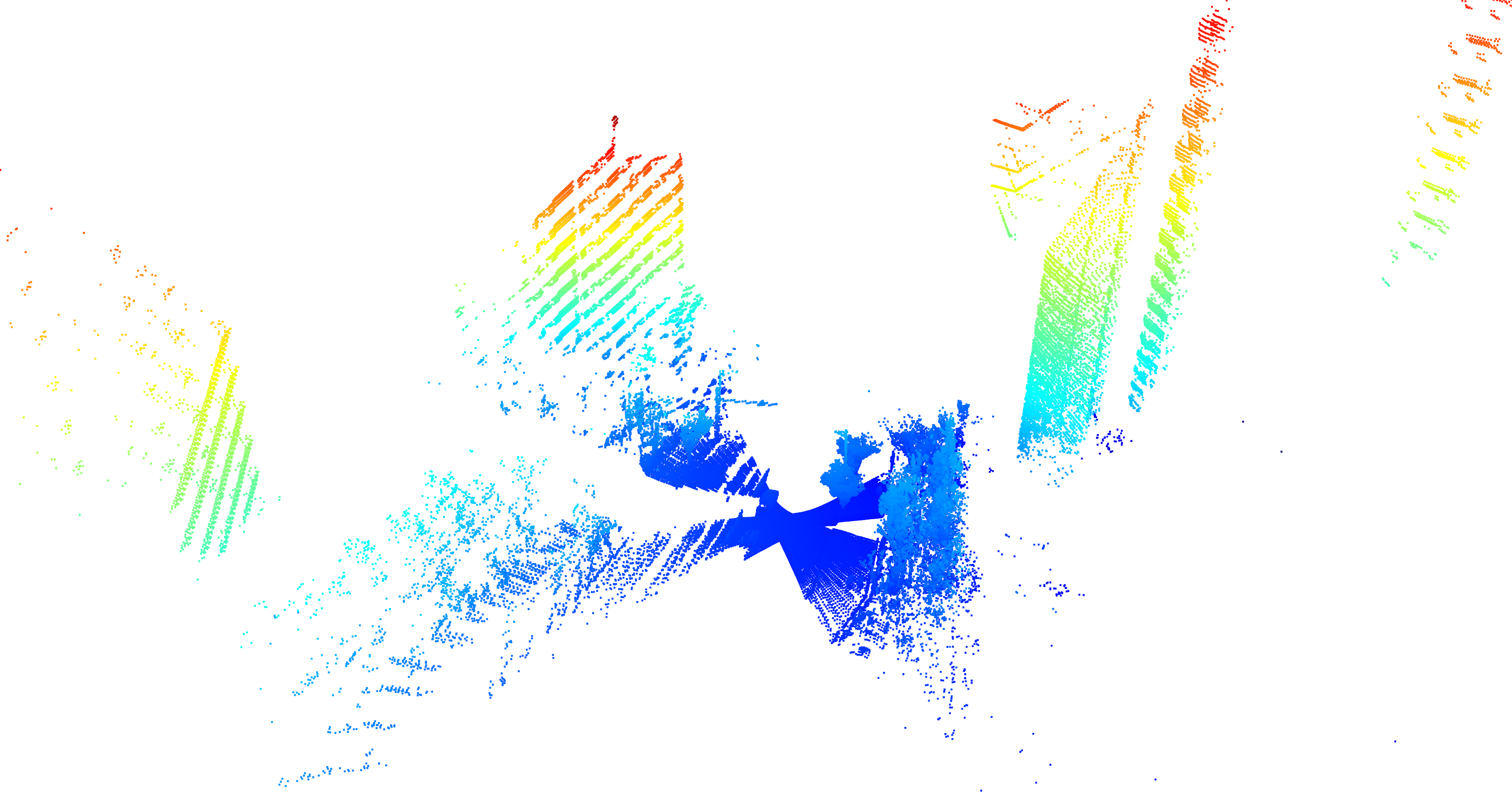}
    \end{minipage}
    \hfill
    \begin{minipage}[t]{0.349\textwidth}
        \includegraphics[width=\linewidth, height=0.12\textheight]{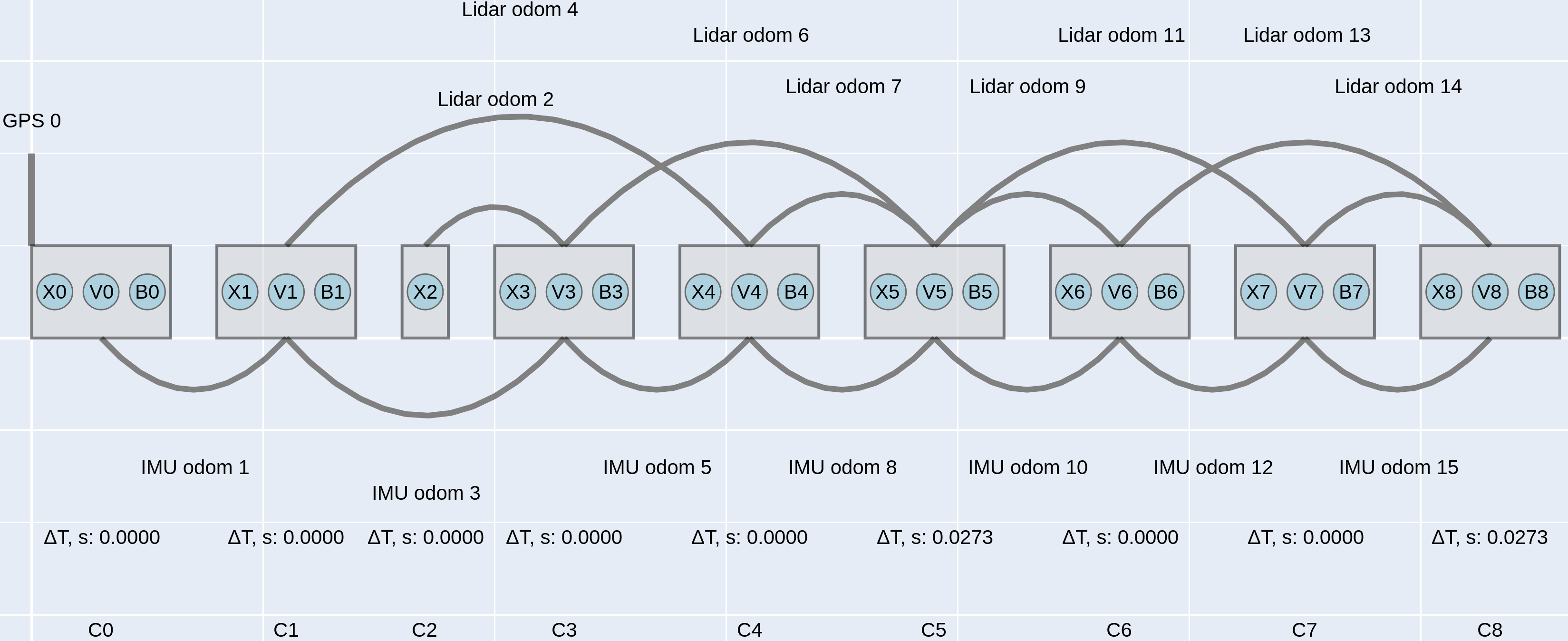}
        \subcaption{Ours}
    \end{minipage}

    \caption{For two measurement sequences, two point cloud maps (middle) were created from factor graphs generated using the Base scenario (a). The factor graphs generated by the proposed algorithm are shown on the right (b). The rectangles illustrate the clusters \(C_i\) with the corresponding vertices. Our approach reduced the number of vertices by \textbf{24.2\%} for the first sequence and by \textbf{33.3\%} for the second one while maintaining the quality of the map. The difference between the point cloud maps generated by the base scenario and our approach is indistinguishable and has not been visualized.}
    \label{fig:four_images}
\end{figure*}

Localization and mapping are crucial for autonomous systems, enabling precise navigation and environmental interaction. Modern robots fuse data from LiDARs, radars, cameras, IMUs, and GPS to build accurate maps. Factor graph~\cite{factor_graph} offers an efficient approach for multimodal sensor fusion, using probabilistic inference to estimate the Maximum a Posteriori (MAP) solution. Unlike Bayesian filters, it performs full batch optimization and efficiently handles loop closures, making it more suitable for mapping~\cite{graphs_vs_kalman}. However, factor graph-based SLAM faces challenges due to asynchronous sensor data and the discrete structure of the graph model. Sensors operate at non-uniform intervals, and key constraints like GPS and landmark detections are non-periodic due to signal loss or environmental factors. In online SLAM, future observations are unknown, while offline analysis of large datasets is memory-intensive. Accurate state estimation requires factors to introduce the corresponding variables at specific times. However, this may create a disconnected graph, isolating nodes and excluding measurements from the tightly coupled optimization process, leading to suboptimal results. This work tackles the challenge of handling asynchronous measurements to maximize information utilization in the optimization process while ensuring graph connectivity for globally consistent state estimation.

Sensor measurements that form the graph factors arrive at varying rates with different timestamps, requiring the creation of new graph variables. The order of measurements is not always periodic and known due to signal loss, sensor failures, or processing delays. In this work we assume that all measurements involved in factors have the same time axis: for time moments \( t_1, t_2, \dots, t_T \) all measurements \( x_1(t), x_2(t), \dots, x_n(t) \) are defined on the same time axis \( t \in \{ t_1, \dots, t_T \} \). These issues may result in an \textit{unconnected factor graph} containing isolated variables, which can be defragmented into multiple separate graphs.

A factor graph in the SLAM problem is a graphical representation of a probabilistic inference model. A \textit{Connected Factor Graph} is a bipartite graph \( G = (V, F, E) \), where \( V = \{v_1, \dots, v_n\} \) is the set of variable nodes, \( F = \{f_1, \dots, f_m\} \) is the set of factor nodes, \(E=\{e_1, \dots, e_k\}\) is the set of edges connecting \(v_i\) and \(f_j\).  An edge \( e_{ij} \in E \) exists iff factor \( f_i \) (a function) depends on the state variable \( v_j \) (a parameter).

A joint density function with a factorized representation over a set of variables $\Theta = [\theta_1, \theta_2, \dots, \theta_n]$ might be represented as: 
\begin{equation}
    \label{eq:inference}
    P(\Theta) \propto \prod_i \underbrace{\phi_i(\theta_i)}_{\text{Unary factor}} \prod_{i < j} \underbrace{\psi_{ij}(\theta_i, \theta_j)}_{\text{Pairwise factor}}
\end{equation}

Maximum a posteriori estimation over this function can be efficiently calculated under the assumption of the Gaussian model for the measurement noise. The likelihood maximization can be formulated as a non-linear least squares minimization problem~\cite{square_root_sam} which linearized version  \(A\delta\text{ = }b\) can be solved incrementally by updating the vector of residuals \(\delta\). The coefficients matrix \(A\) is a block matrix with Jacobians which represents the connections in the corresponding factor graph. To ensure that all variables are tightly coupled in the optimization process, a factor graph should be connected and its matrix \(A\) should be irreducible: \( A \in \mathbb{R}^{m \times n} \) is irreducible if there exist no permutation matrices \( P \) and \( Q \) such that \( PAQ = \begin{bmatrix} A_{11} & 0 \\ A_{12} & A_{22} \end{bmatrix} \), where \( 0 \) is a non-trivial zero block. If a system can not be decomposed into multiple independent sub-systems this ensures that the variables influence each other during the optimization process.

To solve the \textit{unconnected factor graph} problem caused by asynchronous measurements, we introduce a novel method for incremental map construction by formulating a \textit{Connected Factor Graph} that integrates all available sensor measurements, thereby ensuring connectivity and facilitating a tightly coupled optimization process. Our approach generates all potential configurations of factor graphs and selects the sub-optimal one based on the predefined evaluation criteria. This enables the selection of the graph topology that best represents the local map according to the specified metrics. Consequently, the entire map is composed of a combination of sub-optimal local maps.

The contribution of this paper is as follows:
\begin{enumerate}
    \item To address graph connectivity problem arising from asynchronous, multi-rate measurements and to fully exploit available sensor data, we introduce a method that incrementally constructs sub-optimal factor graphs, selecting their topology based on external evaluation criteria.

    \item The proposed approach was validated on a real-world dataset against a canonical factor graph construction method, achieving up to 41.2\% reduction in graph size with no visually distinguishable loss in map quality.
\end{enumerate}

%% file: chapters/related_work.tex
\section{Related Work}
\label{sec:related_work}

\begin{figure}[t]
	\centering
	\includegraphics[width=1.0\columnwidth]{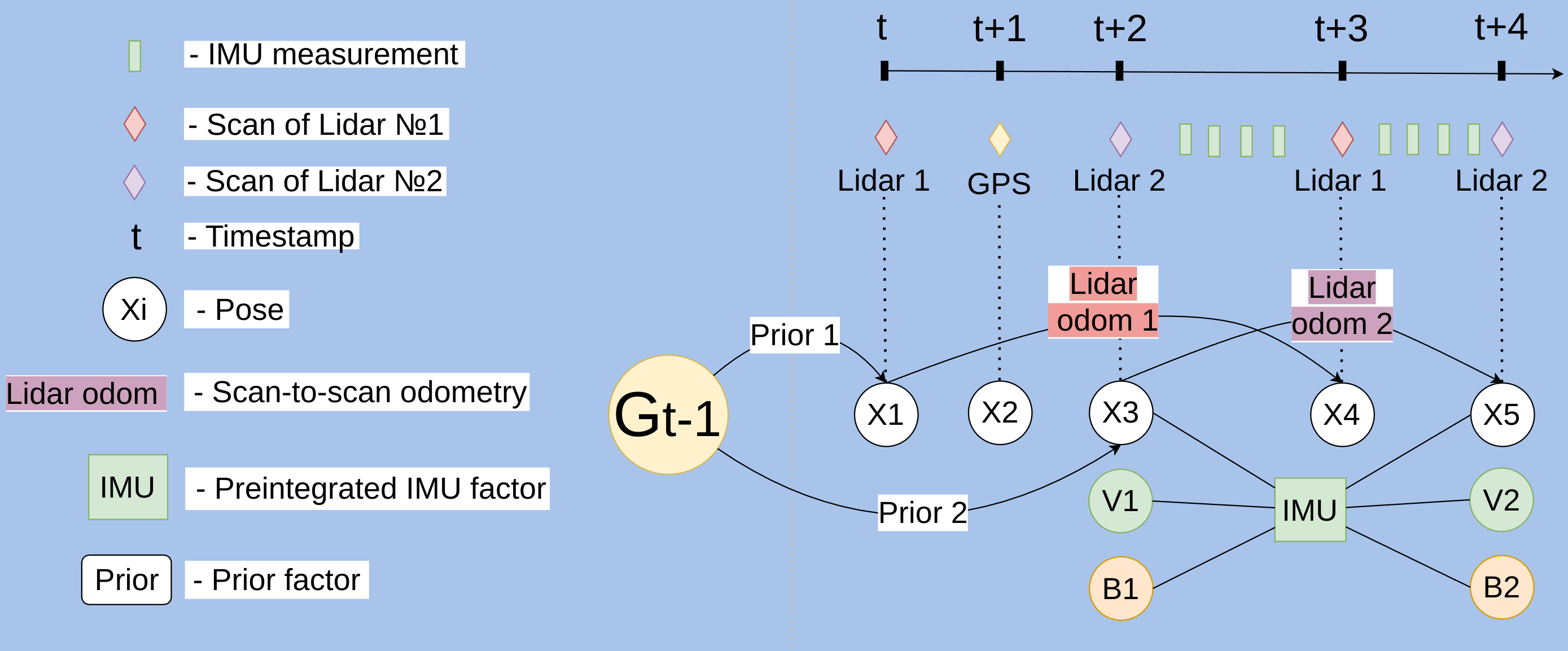}
	\caption{
        A measurement sequence with 4 scans acquired from 2 LiDARs, GPS and IMU data, and the factor graph with poses, velocities, and IMU biases to be estimated. $G_{t-1}$ is a factor graph for the time history $[t_0, ...,  t \text{-} 1]$, with prior factors connecting the poses derived from LiDAR scans. The vertex $X_2$ is not connected with others, and the associated GPS factor does not influence the estimation of all variables.
	}
	\label{fig:problem}
\end{figure}

Sensors such as Inertial Measurement Units (IMUs), fiber-optic gyroscopes, and wheel encoders typically produce measurements at high frequencies, while other sensors, including GPS, cameras, and LiDARs, operate at lower sampling rates. The factors within the system necessitate the introduction of new variables, such as poses, velocities, or IMU biases, which are incorporated into the system of equations. However, these estimated variables (graph nodes) correspond to distinct time instants and do not always form a path in the graph between each other. There are several approaches in the robotics area addressing the challenge of consistent discrete state estimation for continuous dynamics and trying to maximize the information utilization from the sensors.

\subsection{High-frequency measurements integration}

High-frequency measurements can be pre-integrated between successive lower-frequency sensor measurements. This approach reduces the number of variables in the factor graph by summarizing the effect of multiple sensor readings into a single factor, connecting the states at the times of the lower-rate measurements. Classical IMU pre-integration technique~\cite{imu_preint}  discusses this method in detail, highlighting its effectiveness in combining IMU data with visual observations. Although it avoids the estimation of parameters for every reading it demands additional evaluation of velocities, gyroscope, and accelerometer biases in the connected states. As illustrated in Fig.~\ref{fig:problem}, in some datasets (including the one used in this paper) there are measurement sequences with low-rate measurements following each other without IMU readings in between, which limits the use of the pre-integration technique. 

\subsection{Continuous-time trajectory representation}

Modeling the robot's trajectory as a continuous-time function enables querying states at any time, accommodating measurements from sensors with varying sampling rates. This approach integrates asynchronous data by evaluating the continuous trajectory at measurement time moments. Continuous-time spline-based methods~\cite{b_spline_1, b_spline_2, b_spline_3} within factor-graph optimization have been extensively studied, demonstrating accurate pose estimations. These methods represent trajectories as piece-wise polynomial B-spline curves, optimizing control points and its coefficients. However, accurate spline fitting depends on the proper selection of control points, including their distribution and quantity. In practice, uniform distribution is commonly used~\cite{b_spline_4}. Poorly distributed control points can lead to inconsistent trajectory representations, as poses are sampled from the spline curve. Notably, spline-based methods do not address the factor graph connectivity issue described earlier, as measurement timestamps may not align with control points.

\subsection{Measurements interpolation}

Another approach~\cite{measurement_interpolation} to solve an asynchronous timestamps problem is the interpolation of raw measurements. Having a high-frequency sensor (radio or sonar) that measures the distance to the beacons the authors showed that the interpolated range data may improve the quality of estimation without losing computational efficiency. However, examples of real data where the linear interpolation introduces significant measurement error have also been demonstrated. Thus, it is hardly applicable for low-rate sensors (GPS, cameras), especially in high dynamic conditions.

\subsection{Simultaneous filtering \& smoothing}

Filtering and smoothing techniques work in parallel, with filtering using high-frequency measurements for real-time state estimation and smoothing refining variables through loop closures. Filtering handles asynchronous data efficiently while smoothing ensures global consistency. PTAM~\cite{ptam} decouples camera tracking from mapping in monocular SLAM, performing localization and map refinement via bundle adjustment~\cite{bundle_adjustment} concurrently. Unlike navigation problems requiring joint pose and velocity estimation, PTAM relies on repeated camera re-localization relative to the map. KinectFusion~\cite{kinect_fusion} adopts a similar parallelization strategy. While effective for visual tracking, this separation lacks probabilistic integration of inertial data. To address this, the Concurrent Filtering and Smoothing technique~\cite{CFS} was introduced combining filtering, batch smoothing for loop closures and "separators" to link interrelated states. Filtering predicts states up to the next low-frequency measurement (e.g., GPS) and updates them in constant time, while batch optimization ensures global trajectory consistency, requiring at least linear time relative to the number of variables. However, synchronization between smoother and filter introduces discrete updates at separator nodes, causing abrupt state changes, as shown by error and trajectory plots in the study.

\subsection{Factor graph sparsification}

Graph compression can be achieved by removing outdated or redundant states. Different sparsification techniques marginalize variables while maintaining global consistency and sparsity patterns. The process involves two steps: marginalizing selected variables, usually via the Schur complement, and approximating the resulting dense multivariate Normal distribution for factorization with a sparsified version, represented by new re-linearizable factors. Methods like \textit{Generic Node Removal (GNR)}~\cite{GNR} and \textit{Information-Theoretic Compression}~\cite{map_entropy} use Chow-Liu trees to minimize the \textit{Kullback-Leibler Divergence (KLD)} in tree-structured approximations. The latter refines sparsification by selecting the most informative laser scan nodes based on \textit{conditional entropy}, removing redundant nodes while adding new ones only when they provide significant information gain. This aligns with our approach of analyzing the environment before finalizing the solution. However, the authors are more focused on restricting the factor graph size by rejecting non-informative laser scans than on maintaining tightly coupled state estimation with maximum information utilization. Beyond tree-based structures, methods like \textit{Nonlinear Factor Recovery (NFR)}~\cite{NFR} and \textit{Factor Descent}~\cite{factor_decent_sparcification} iteratively optimize factors by minimizing the \textit{KLD} between \( p(x) \) and \( q(x) \). While these methods excel in graph compression, they focus on recovering nonlinear factors representing marginal distributions rather than solving the factor graph connectivity problem caused by asynchronous measurements. Thus, the direct comparison with the proposed method is inappropriate.

%% file: chapters/methodology.tex
\section{Methodology}
\label{sec:methodology}

SLAM enables concurrent robot localization and environmental mapping. Real-time state estimation for navigation requires approximately constant computational time, while high-quality long-range maps demand significant resources and are usually generated offline. This paper follows the same paradigm. There are several reasons for that. Factor graphs outperform traditional filtering methods by enabling full batch optimization, and incorporating loop closures for global consistency. However, this requires the full system of equations to be loaded into memory. Different relaxation techniques~\cite{se_sync, certifiable_slam} can provide certifiably correct global solutions avoiding local minima but demand to solve the dual optimization problem. State-of-the-art methods~\cite{semantic_segm_slam, depth_slam} often use neural networks for classification, segmentation, or depth estimation. Although the methods are GPU-parallelizable, they require careful subsequent synchronization. Another challenge of long-term mapping is memory limitation, which prevents measurements from being loaded into memory for arbitrarily large datasets. This requires iterative processing of measurements, which is efficiently implemented in the ModuSLAM~\cite{moduslam} framework.

The classical graph-based SLAM pipeline involves: \textbf{1.} processing measurements, \textbf{2.} associating them with existing variables or creating new ones, and \textbf{3.} adding corresponding factors to the graph. We propose accumulating a sequence of measurements until a user-defined criterion is met. This criterion is application-dependent; for incremental mapping, it is beneficial to collect sufficient data for a local sub-map. In our experiments, we accumulate and register two LiDAR scans (from left and right sensors) alongside other measurements like GPS and IMU, as shown in Fig.~\ref{fig:problem}.

\subsection{Definitions}
\label{sec:definitions}

\begin{itemize}

	\item Measurement sequence $\mathcal{M} = [m_1, ..., m_N]$ is a sequence of $N$ sensor measurements sorted by timestamp and used to create factors.

	\item Measurement cluster $C = [m_1, ..., m_K], m_k \in \mathcal{M}$ is a sorted sequence of sensor measurements grouped together.

	\item Measurement cluster timestamp $t^C$ is a timestamp of the median element $m \in C$.
	
	\item Measurement cluster time range $\Delta{T^C} = t_K - t_1$ is a time difference between the 1-st and the last measurement in the cluster.

\end{itemize}

\subsection{Algorithm.}
\label{sec:graph_candidate}

\begin{algorithm}
	\caption{Sub-optimal factor graph generation}
	\label{alg:algorithm}
	\begin{algorithmic}[1]
	
	\State \textbf{Input:} \hspace{0.5em} \begin{tabular}[t]{@{}l@{}}
		Factor Graph $G_{t-1}$, \\
		measurements $\mathcal{M} = \{m_1, m_2, \dots, m_N\}$
	\end{tabular}

	\State \textbf{Output:} factor graph $G_t$
	
	\State \textbf{Step 1:} Split all measurements into core $\mathcal{M}^C$ and continuous $\mathcal{M}^{I}$ measurements.
	
	\State \textbf{Step 2:} Create sequence $\mathcal{C}^N = \{C_1, C_2, \dots, C_N\}$ of \(N\) clusters, where each  $C_i$ includes $m_i \in \mathcal{M}^C$.

	\State \textbf{Step 3:} Generate all possible combinations by merging the adjacent clusters $C_i, C_{i+1}$.
	
	\State \textbf{Step 4:} If \(|\mathcal{M}^{I}| \neq 0\) add connections between clusters:
		\ForAll{combinations of measurement clusters}
			\State Generate all valid connections between $[C_i, C_{j}]$ using measurements from $\mathcal{M}^{I}$.
		\EndFor
	
	\State \textbf{Step 5:} Filter \textit{unconnected factor graphs}, solve \& evaluate the remaining ones based on the defined metrics.
	
	\end{algorithmic}
\end{algorithm}

The proposed \textbf{Algorithm}~\ref{alg:algorithm} processes a sequence of measurements $\mathcal{M}$ and the previous factor graph $G_{t\text{-}1}$ and outputs a sub-optimal (up to the evaluation scheme) factor graph $G_{t}$ as outlined below:

\textbf{Measurements categorization (step 1)}:  
the input measurements are categorized into two types: \textit{core measurements} $\mathcal{M}^C$ and \textit{continuous measurements} $\mathcal{M}^I$ (if available). Core measurements, such as LiDAR scans, GPS data, or detected landmarks, are essential for constructing a factor graph and a map. High-rate measurements, such as IMU accelerations \& angular velocities, wheel encoder outputs, or fiber optic gyroscope (FOG) angular velocities, named ``\textit{continuous measurements}" are utilized to generate pre-integrated factors. These factors serve to connect the variables derived from core measurements.

\textbf{Clusters initialization (step 2)}: \(N\) measurement clusters are created, with each core measurement assigned to its own cluster. Each cluster obtains time properties based on the timestamps of the core measurements it contains.

\textbf{Generation of clusters' combinations (step 3)}:  
all possible combinations of clusters are generated by merging only the adjacent (neighboring) clusters. For instance, given three measurements with timestamps \( m_{t_1}, m_{t_2}, m_{t_3} \), the valid cluster combinations are: \([m_{t_1}]\), \([m_{t_2}]\), \([m_{t_3}]\); \([m_{t_1}, m_{t_2}], [m_{t_3}]\); \([m_{t_1}], [m_{t_2}, m_{t_3}]\), and \([m_{t_1}, m_{t_2}, m_{t_3}]\). The combination \([m_{t_1}, m_{t_3}]\) is invalid due to the absence of \( m_{t_2} \) between \( m_{t_1} \) and \( m_{t_3} \). Loops defined as binary connections between the same vertices are rejected. The total number of unique combinations is at most \( 2^{N-1} \) and is approximately constant for a fixed amount of sensors and a fixed time window. The detailed derivations are provided in Appendix~\ref{sec:appendix}.

\textbf{Generation of connections' combinations (step 4)}:  
for each valid cluster combination, binary connections are generated if at least one continuous measurement $m \in \mathcal{M}^I$ is present between clusters $C_i, C_{i+1}$. To maintain probabilistic consistency and measurement independence, which is crucial for Eq.~\ref{eq:inference}, each $m \in \mathcal{M}^I$ is used only once, and no intersections with other connections are allowed. Each combination of \( K \) measurement clusters yields at most \( 2^{K-2} \) valid combinations of connections.

\textbf{Factor Graph Evaluation (step 5)}:  
all unconnected factor graphs are filtered out. The systems of equations of the remaining ones are solved in parallel. The resulting local maps are evaluated based on predefined metrics. The candidate \( G_{\text{best}} \) that yields the best map approximation is selected.

\begin{figure}[t]
	\centering
	\includegraphics[width=1.0\columnwidth]{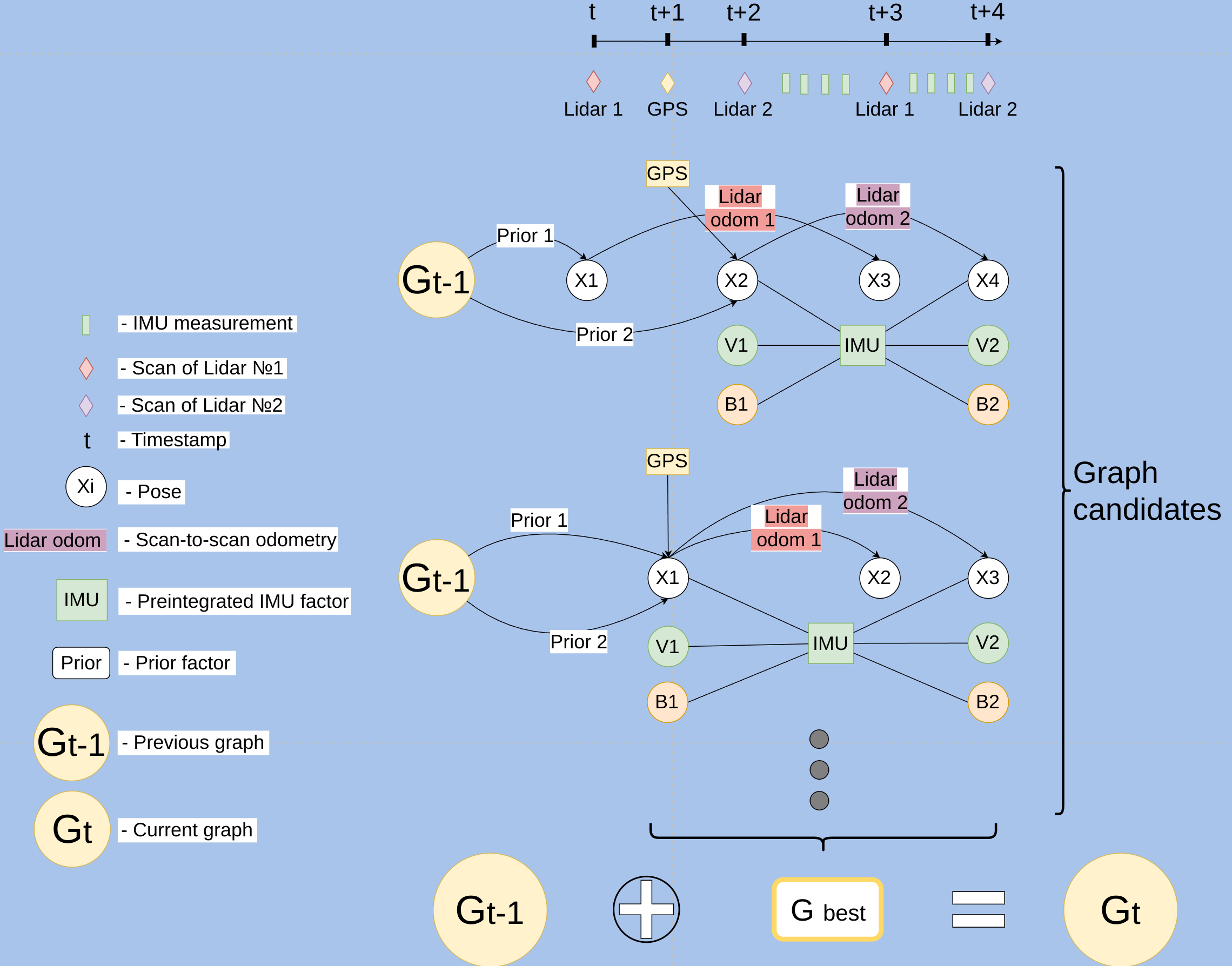}
	\caption{Different factor graph candidates generated by the proposed algorithm for the same sequence of measurements. The resulting graph \( G_{t} \) is a combination \( \oplus \) of the previous graph \( G_{t-1} \) and the factors form the chosen candidate \( G_{\text{best}} \). The 1-st candidate clusters the GPS measurement and the scan from LiDAR 2, while the 2-nd one combines it with two scans from both LiDARs.}
	\label{fig:incremental_graph}
\end{figure}

The proposed algorithm generates multiple factor graphs, each representing a map hypothesis, by clustering neighboring measurements and introducing estimated variables based on the associated factors. Fig.~\ref{fig:incremental_graph} demonstrates different solutions to the graph connectivity problem depicted in Fig.~\ref{fig:problem}. The first graph
candidate combines the GPS measurement and the scan from LiDAR 2. The second one combines the GPS measurement and two scans from both LiDARs into the same cluster. The resulting graphs are connected because there is a path \(P\) between any \(V_i \in G_k\) and a previous graph \(G_{t-1}\). The GPS factor influences the estimation of other variables in the system. A particular graph candidate, named \( G_{\text{best}} \), is chosen by the evaluation scheme, and its factors are added to the previous graph \( G_{t-1} \) to form the resulting \( G_t \). At high speeds or during rapid accelerations, the merging procedure induces displacement errors dependent on time gaps and robot dynamics, potentially degrading map quality. To mitigate this, an external metric quantifies inconsistencies, discarding hypotheses with excessive merging. Conversely, during slow movement or stationarity, merging is feasible if it preserves map accuracy, reducing the number of estimated variables and compressing the factor graph. The algorithm ensures that the resulting factor graph remains connected, guaranteeing that all variables are tightly coupled and contribute to the estimation process. Fig.~\ref{fig:four_images} illustrates an example where the proposed approach achieves compression rates of 24.2\% and 33.3\%, in contrast to the \textit{base} scenario described in Section~\ref{sec:experiments}, for two different measurement sequences while maintaining approximately the same map quality. Base and our approaches created visually indistinguishable point cloud maps thus, there is no sence to ilustrate them both.

\subsection{Evaluation Metric}
The evaluation scheme used to assess the quality of graph candidates must be representative and accurately reflect reality. The choice of a specific metric depends on the type of map and the intended application. In our work, which focuses on LiDAR point clouds, the \textit{Mutually Orthogonal Metric} (MOM)~\cite{mom} is a suitable candidate for evaluating point cloud alignment in urban environments. The MOM operates by extracting points that lie on 3 mutually orthogonal surfaces. The directions of these surfaces form an orthogonal basis for decomposing the translation part of the Relative Pose Error (RPE), which measures the drift between the reference pose \( T_{\text{ref}} \) and the estimated pose \( T_{\text{est}} \):
\begin{equation}
    \label{eq:rpe}
    \begin{aligned}
        &E = (T^{-1}_{\text{ref}} T_{\text{ref}})^{-1} (T^{-1}_{\text{est}} T_{\text{est}}) \in \text{SE}(3), \\
        &E_{\text{trans}} = \lVert \text{trans}(E) \rVert_2.
    \end{aligned}
\end{equation}

For each subset of points along a basis direction, the mean plane variance of the map is:

\begin{equation}
    V(P) = \frac{1}{|P|} \sum_{k=1}^{|P|} \lambda_{\text{min}}(\Sigma(W(p_k)))
\end{equation}
where \( W(p_k) \) is vicinity of every point \(p_k\) in the aggregated map \(P\). The covariance matrix \( \Sigma(W(p_k)) \) captures the spatial distribution, and \( \lambda_{\text{min}}(\Sigma(W(p_k))) \) is its smallest eigenvalue, representing the variance orthogonal to the best-fit local plane. The MOM value, defined in Eq.~\ref{eq:local_mom}, is the average of the three mean plane variances along the basis directions. It is evaluated on a local sub-map \(P\), which is the intersection between the existing point cloud and a new point cloud from recent poses. 

\begin{equation}
    \label{eq:local_mom}
        E^{MOM}_P = \frac{1}{3}(V(P)_x + V(P)_y + V(P)_z), 
\end{equation}

The total trajectory metric is the sum of the MOM values across 
\(N\) local sub-maps:

\begin{equation}
    \label{eq:total_mom}
        E^{MOM} = \sum_{i=1}^{N} E^{MOM}_{P_i}.
\end{equation}

This approach provides a robust measure of alignment quality in environments with mutually orthogonal structures. The primary advantage of this metric is its ability to leverage the geometric properties of such structures, which are common in urban outdoor scenes (e.g., planar roads, perpendicular building facades, fences, and road signs). However, in environments lacking such structures, such as tunnels, fields, or highways, the MOM may not be applicable. Additionally, this metric us purely based on the decomposing of the translation part of the RPE and does not take into account the rotation part.

%% file: chapters/experiments.tex
\section{Experiment evaluation}
\label{sec:experiments}

\subsection{Experimental Setup}
\label{sec:setup}

The experiments utilize the KAIST Urban dataset~\cite{kaist}, which features diverse urban environments: highways, streets, and buildings with varying car dynamics. The sensor setup consists of VRS GPS with a status 4 (fixed) solution, left \& right roof-mounted LiDARs, and an IMU. A voxel-based scan matching algorithm~\cite{kiss_icp} computes transformation \(T_{i,i+1} \in SE(3)\) between consecutive left-to-left or right-to-right LiDAR scans without any environment-specific fine-tuning. ModuSLAM~\cite{moduslam} uses IMU preintegration factors and the Levenberg-Marquardt solver implemented in GTSAM~\cite{gtsam}. GPS fixed status is infrequent, primarily occurring on highways where the MOM metric fails due to the lack of orthogonal objects in LiDAR scans. Another obstacle is the LiDARs' \( \text{45}^{\circ} \) mounting angle that yields sparse point clouds, mainly capturing the road and sky, hindering complete map reconstruction. For a comprehensive evaluation, the dataset is segmented by versatile motion scenes: stationary (0 km/h), acceleration (\(\sim\) 0-20 km/h), deceleration (\(\sim\) 22-0 km/h), and constant-speed (\(\sim\) 18-22 km/h), ensuring fixed GPS status and orthogonal objects. Each data sequence lasts \(\sim\) 7-10 seconds. All the generated graph candidates are categorized into four scenarios:

\begin{enumerate}
    \item \textbf{Base:}  
    a conventional scenario with variables estimated exactly at the time instances of the corresponding measurements without any clustering, and sequentially connected by pre-integrated IMU factors (if available).

    \item \textbf{Minimum Time Shift:}  
    all factors and variables are created such that the minimum time shift for all measurements is achieved:
    \begin{equation}
    \label{eq:timeshift}
        E = T_{\text{time range}} + T_{\Delta \text{IMU}} = \sum_{i=1}^{N} \Delta t_i + \sum_{j=1}^{K} \Delta t_j,
    \end{equation}
    where \( T_{\text{time range}} = \sum_{i=1}^{N} \Delta T^C_i \) represents the sum of measurement cluster time ranges. The IMU factor is a function \( f(X, V, B, \Sigma, [m_1, \dots, m_N]) \) of poses, velocities, biases, covariance, and IMU measurements. The total time shift for all IMU preintegration factors is \( T_{\Delta \text{IMU}} = \sum_{j=1}^{K} \Delta t_j \), where \( \Delta t_j = t_{m_1} - t^{C_i} \) is the time shift for the \( j \)-th IMU factor.

    \item \textbf{Minimum Solver Error:}  
    the scenario with the least Mahalanobis distance for all factors used in a graph candidate is given by:
    \begin{equation}
    \label{eq:mahalanobis}
        E = \sum_{i=1}^{N} \left\| \mathbf{h}_i(\mathbf{x}) - \mathbf{z}_i \right\|^2_{\mathbf{\Sigma}_i},
    \end{equation}
    where \( N \) is the total number of factors, \( \mathbf{h}_i(\mathbf{x}) \) is the measurement model for the \( i \)-th factor, \( \mathbf{z}_i \) is the observed measurement, and \( \mathbf{\Sigma}_i \) is the measurement noise covariance matrix.

    \item \textbf{MOM-Based:}  
    A graph candidate that produces a local point cloud map with the least MOM value.
\end{enumerate}

\subsection{Results}
\label{sec:results}

\begin{table*}[t]\centering
    \caption{Evaluation results for 4 different scenarios on 3 datasets for 4 dynamic states. The best results are highlighted in green, the second-best in light green, and the worst in red.}
    \label{tab:results}
    \tiny
    \renewcommand{\arraystretch}{1.3}
    \resizebox{\textwidth}{!}{
    \begin{tabular}{|l|r|r|r|r||r|r|r|r||r|r|r|r|}
    \hline
    & \multicolumn{4}{c||}{\textbf{Urban-26}} & \multicolumn{4}{c||}{\textbf{Urban-33}} & \multicolumn{4}{c|}{\textbf{Urban-38}} \\
    \cline{2-13}
    & \multicolumn{12}{c|}{\textbf{Zero-speed}} \\
    \hline
    & \textbf{Base} & \textbf{\textcolor{blue}{Ours}} & \textbf{Min. time shift} & \textbf{Min. solver error} 
    & \textbf{Base} & \textbf{\textcolor{blue}{Ours}} & \textbf{Min. time shift} & \textbf{Min. solver error} 
    & \textbf{Base} & \textbf{\textcolor{blue}{Ours}} & \textbf{Min. time shift} & \textbf{Min. solver error} \\
    \hline
    Total time shift, sec & 1.0105 & 3.5686 & 0.5112 & 0.8188 & 0.7318 & 1.0892 & 0.5566 & 0.7153 & 1.0167 & 1.1664 & 0.5298 & 0.6084 \\
    Solver error & 0.0224 & 0.0167 & 0.0103 & 0.0097 & 0.0241 & 0.0227 & 0.0077 & 0.0085 & 0.0319 & 0.0263 & 0.0181 & 0.0177 \\
    RPE, m & \cellcolor[HTML]{6aa84f}\textbf{0.001} & \cellcolor[HTML]{b6d7a8}\textbf{0.004} & \cellcolor[HTML]{ea9999}0.013 & \cellcolor[HTML]{ea9999}0.013 & \cellcolor[HTML]{6aa84f}\textbf{0.008} & \cellcolor[HTML]{b6d7a8}\textbf{0.009} & \cellcolor[HTML]{ea9999}0.014 & \cellcolor[HTML]{ea9999}0.014 & \cellcolor[HTML]{6aa84f}\textbf{0.001} & \cellcolor[HTML]{b6d7a8}\textbf{0.005} & \cellcolor[HTML]{ea9999}\textbf{0.016} & 0.015 \\
    RPE, deg & \cellcolor[HTML]{6aa84f}\textbf{0.009} & \cellcolor[HTML]{b6d7a8}\textbf{0.013} & 0.117 & \cellcolor[HTML]{ea9999}0.169 & \cellcolor[HTML]{6aa84f}\textbf{0.037} & \cellcolor[HTML]{b6d7a8}\textbf{0.038} & \cellcolor[HTML]{ea9999}0.058 & 0.055 & \cellcolor[HTML]{6aa84f}\textbf{0.011} & \cellcolor[HTML]{b6d7a8}\textbf{0.021} & \cellcolor[HTML]{ea9999}\textbf{0.204} & 0.179 \\
    clusters & 211 & 165 & 210 & 202 & 213 & 175 & 212 & 208 & 211 & 194 & 211 & 207 \\
    vertices & 632 & 435 & 432 & 412 & 637 & 463 & 514 & 474 & 633 & 528 & 435 & 427 \\
    factors & 417 & 343 & 318 & 312 & 383 & 335 & 323 & 320 & 417 & 374 & 318 & 316 \\
    compression, \% & 0.00 & \textbf{31.17} & \cellcolor[HTML]{b6d7a8}\textbf{31.65} & \cellcolor[HTML]{6aa84f}\textbf{34.81} & 0.00 & \cellcolor[HTML]{6aa84f}\textbf{27.32} & 19.31 & \cellcolor[HTML]{b6d7a8}\textbf{25.59} & 0.00 & \textbf{16.59} & \cellcolor[HTML]{b6d7a8}\textbf{31.28} & \cellcolor[HTML]{6aa84f}\textbf{32.54} \\
    \hline
    & \multicolumn{12}{c|}{\textbf{Acceleration}} \\
    \hline
    Total time shift & 0.7064 & 3.5525 & 0.3684 & 0.5944 & 0.3161 & 0.4217 & 0.2084 & 0.2688 & 0.5062 & 0.7803 & 0.2493 & 0.2616 \\
    Solver error & 6.0100 & 7.2389 & 3.9376 & 3.8399 & 2.5002 & 2.4815 & 1.2488 & 0.8144 & 0.6140 & 0.6330 & 0.2819 & 0.2655 \\
    RPE, m & \cellcolor[HTML]{6aa84f}\textbf{0.087} & \cellcolor[HTML]{b6d7a8}\textbf{0.229} & \cellcolor[HTML]{ea9999}1.870 & 0.763 & \cellcolor[HTML]{6aa84f}\textbf{0.081} & \cellcolor[HTML]{b6d7a8}\textbf{0.098} & 0.219 & \cellcolor[HTML]{ea9999}0.577 & \cellcolor[HTML]{6aa84f}\textbf{0.023} & \cellcolor[HTML]{b6d7a8}\textbf{0.050} & \cellcolor[HTML]{ea9999}\textbf{0.421} & 0.302 \\
    RPE, deg & \cellcolor[HTML]{6aa84f}\textbf{0.544} & \cellcolor[HTML]{b6d7a8}\textbf{0.792} & \cellcolor[HTML]{ea9999}10.530 & 9.525 & \cellcolor[HTML]{6aa84f}\textbf{0.410} & \cellcolor[HTML]{b6d7a8}\textbf{1.276} & 8.564 & \cellcolor[HTML]{ea9999}27.263 & \cellcolor[HTML]{6aa84f}\textbf{0.209} & \cellcolor[HTML]{b6d7a8}\textbf{0.329} & 15.659 & \cellcolor[HTML]{ea9999}\textbf{16.183} \\
    clusters & 148 & 99 & 147 & 141 & 87 & 68 & 87 & 84 & 106 & 91 & 106 & 105 \\
    vertices & 444 & 261 & 301 & 287 & 259 & 176 & 197 & 172 & 318 & 243 & 220 & 219 \\
    factors & 292 & 226 & 222 & 217 & 163 & 136 & 132 & 129 & 208 & 179 & 159 & 159 \\
    compression, \% & 0.00 & \cellcolor[HTML]{6aa84f}\textbf{41.22} & 32.21 & \cellcolor[HTML]{b6d7a8}\textbf{35.36} & 0.00 & \cellcolor[HTML]{b6d7a8}\textbf{32.05} & 23.94 & \cellcolor[HTML]{6aa84f}\textbf{33.59} & 0.00 & \textbf{23.58} & \cellcolor[HTML]{b6d7a8}\textbf{30.82} & \cellcolor[HTML]{6aa84f}\textbf{31.13} \\
    \hline
    & \multicolumn{12}{c|}{\textbf{Constant speed}} \\
    \hline
    Total time shift & 0.7314 & 1.3454 & 0.3794 & 0.3829 & 0.4115 & 2.0657 & 0.2321 & 0.4056 & 0.3801 & 0.6146 & 0.3447 & 0.4684 \\
    Solver error & 3.8407 & 3.999 & 1.3795 & 1.3493 & 3.5212 & 3.9112 & 2.7542 & 0.6130 & 0.7016 & 1.9143 & 0.3866 & 0.2675 \\
    RPE, m & \cellcolor[HTML]{6aa84f}\textbf{0.075} & \cellcolor[HTML]{b6d7a8}\textbf{0.144} & 6.290 & \cellcolor[HTML]{ea9999}\textbf{6.458} & \cellcolor[HTML]{6aa84f}\textbf{0.046} & \cellcolor[HTML]{b6d7a8}\textbf{0.118} & 1.231 & \cellcolor[HTML]{ea9999}3.880 & \cellcolor[HTML]{6aa84f}\textbf{0.148} & \cellcolor[HTML]{b6d7a8}\textbf{0.216} & 0.293 & \cellcolor[HTML]{ea9999}\textbf{0.550} \\
    RPE, deg & \cellcolor[HTML]{6aa84f}\textbf{0.426} & \cellcolor[HTML]{b6d7a8}\textbf{0.852} & 9.263 & \cellcolor[HTML]{ea9999}\textbf{9.287} & \cellcolor[HTML]{6aa84f}\textbf{0.216} & \cellcolor[HTML]{b6d7a8}\textbf{0.329} & \cellcolor[HTML]{ea9999}32.901 & 6.925 & \cellcolor[HTML]{b6d7a8}\textbf{3.075} & \cellcolor[HTML]{6aa84f}\textbf{1.904} & \cellcolor[HTML]{ea9999}\textbf{4.261} & 3.423 \\
    clusters & 150 & 122 & 150 & 149 & 85 & 55 & 85 & 81 & 128 & 112 & 128 & 123 \\
    vertices & 450 & 330 & 308 & 307 & 255 & 151 & 175 & 163 & 380 & 290 & 340 & 263 \\
    factors & 297 & 251 & 226 & 226 & 167 & 131 & 127 & 124 & 213 & 201 & 193 & 187 \\
    compression, \% & 0.00 & \textbf{26.67} & \cellcolor[HTML]{b6d7a8}\textbf{31.56} & \cellcolor[HTML]{6aa84f}\textbf{31.78} & 0.00 & \cellcolor[HTML]{6aa84f}\textbf{40.78} & 31.37 & \cellcolor[HTML]{b6d7a8}\textbf{36.08} & 0.00 & \cellcolor[HTML]{b6d7a8}\textbf{23.68} & 10.53 & \cellcolor[HTML]{6aa84f}\textbf{30.79} \\
    \hline
    & \multicolumn{12}{c|}{\textbf{Deceleration}} \\
    \hline
    Total time shift & 0.5337 & 0.7121 & 0.2625 & 0.5208 & 0.5071 & 0.5944 & 0.3785 & 0.4262 & 0.8970 & 1.2073 & 0.4599 & 0.7287 \\
    Solver error & 4.3972 & 4.3224 & 2.9855 & 2.4820 & 10.8527 & 14.2963 & 7.8400 & 6.5431 & 1.4828 & 1.3317 & 2.5446 & 2.2125 \\
    RPE, m & \cellcolor[HTML]{6aa84f}\textbf{0.041} & \cellcolor[HTML]{b6d7a8}\textbf{0.052} & \cellcolor[HTML]{ea9999}5.657 & 4.056 & \cellcolor[HTML]{b6d7a8}0.344 & \cellcolor[HTML]{6aa84f}\textbf{0.157} & 0.447 & \cellcolor[HTML]{ea9999}9.003 & \cellcolor[HTML]{6aa84f}\textbf{0.031} & \cellcolor[HTML]{b6d7a8}\textbf{0.055} & 0.713 & \cellcolor[HTML]{ea9999}\textbf{3.007} \\
    RPE, deg & \cellcolor[HTML]{6aa84f}\textbf{0.425} & \cellcolor[HTML]{b6d7a8}\textbf{0.535} & \cellcolor[HTML]{ea9999}33.533 & 20.778 & \cellcolor[HTML]{b6d7a8}1.701 & \cellcolor[HTML]{6aa84f}\textbf{1.006} & 4.512 & \cellcolor[HTML]{ea9999}24.021 & \cellcolor[HTML]{6aa84f}\textbf{0.198} & \cellcolor[HTML]{b6d7a8}\textbf{0.346} & \cellcolor[HTML]{ea9999}\textbf{54.664} & 53.580 \\
    clusters & 108 & 78 & 108 & 102 & 147 & 122 & 145 & 139 & 190 & 150 & 190 & 180 \\
    vertices & 324 & 214 & 222 & 206 & 441 & 336 & 351 & 285 & 570 & 396 & 392 & 366 \\
    factors & 214 & 174 & 163 & 158 & 262 & 242 & 220 & 215 & 375 & 310 & 286 & 280 \\
    compression, \% & 0.00 & \cellcolor[HTML]{b6d7a8}\textbf{33.95} & 31.48 & \cellcolor[HTML]{6aa84f}\textbf{36.42} & 0.00 & \cellcolor[HTML]{b6d7a8}\textbf{23.81} & 20.41 & \cellcolor[HTML]{6aa84f}\textbf{35.37} & 0.00 & \textbf{30.53} & \cellcolor[HTML]{b6d7a8}\textbf{31.23} & \cellcolor[HTML]{6aa84f}\textbf{35.79} \\
    \hline
    \end{tabular}
    }
\end{table*}

The experimental results for Urban-26, Urban-33, and Urban-38 datasets are presented in Table~\ref{tab:results}. Since we focus on LiDAR point cloud maps, the most critical aspect is the accuracy of the estimated poses, as the corresponding scans are placed within them. The datasets provide ground-truth trajectories, which are used for evaluation. The authors of the dataset do not recommend using this ground-truth for evaluating the trajectory because it has been generated using the SLAM algorithm. However, the provided point cloud map looks globally consistent and accurate enough to be compared with the one we obtained. For each scenario, we calculate the Root Mean Squared Error (RMSE) of all RPEs defined in Eq.~\ref{eq:rpe} for both the translation (in meters) and rotation (in degrees) parts by comparing them with the provided ground-truth trajectory as follows:
\begin{equation}
    \begin{aligned}
        &RPE^{\text{trans}}_{i,j} = | \text{trans}(E_{ij}) |_2, \\
        &RPE^{\text{rot}}_{i,j} = \left| \text{angle} \left( \log_{SO(3)}(\text{rot}(E_{ij})) \right) \right|_2, \\
        &RMSE = \sqrt{\frac{1}{N} \sum_{i,j} RPE^2_{ij}}
    \end{aligned}
\end{equation}

Here, the ``trans'' and ``rot'' operators extract the translation vector \(t \in \mathbb{R}^3 \) and the rotation matrix \( R \in SO(3) \) from the transformation \( T \in SE(3) \), \(\log{SO(3)}\) is the logarithm map from the Lie group \( SO(3) \) to the Lie algebra \( \mathfrak{so}(3) \) (a skew-symmetric matrix), and the ``angle'' operator maps \( \mathfrak{so}(3) \) to a vector of Euler angles.

``Total time shift" refers to the cumulative time shift over the entire trajectory, computed using Eq.~\ref{eq:timeshift}. ``Solver error" represents the Mahalanobis distance, calculated using Eq.~\ref{eq:mahalanobis} for the factors in the selected graph candidate and accumulated over the entire trajectory. The terms ``Clusters," ``Vertices," and ``Factors" denote the number of measurement clusters, estimated variables, and factors in the graph, respectively. ``Compression" indicates the percentage reduction in the number of factors in the graph compared to the \textbf{base scenario}. Across all datasets, the proposed method yields qualitatively comparable results in terms of trajectory errors (\( RPE^{\text{trans/rot}} \)) while significantly reducing the factor graph size by \(\sim16-41\%\), depending on robot dynamics.

%% file: chapters/discussion.tex
\section{Discussion}
\label{sec:discussion}

\begin{figure}[h]
	\centering
	\includegraphics[width=1.0\columnwidth]{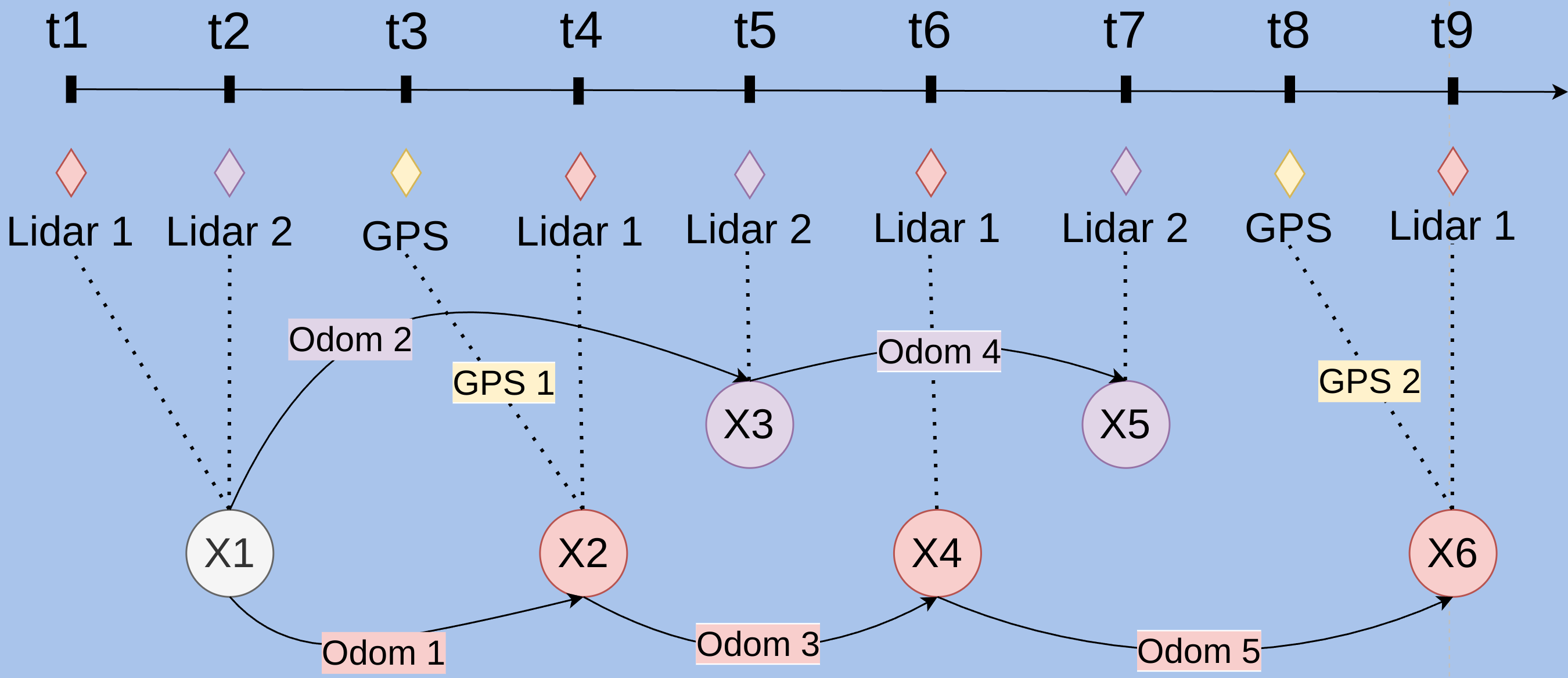}
	\caption{A \textit{connected factor graph} with LiDAR 1 (red) and LiDAR 2 (purple) scan poses illustrates a challenge in the \textit{Minimal Time Shift} and \textit{Minimal Solver Error} scenarios. The variables are loosely coupled, and the GPS factors do not compensate for the drift in odometry measurements obtained from scans of LiDAR 2.}
	\label{fig:error-timeshift-problem}
\end{figure}

As demonstrated in the previous section, the proposed method achieves comparable trajectory accuracy while significantly compressing the factor graph by reducing the number of estimated variables. This improvement stems from two aspects:

1. The pose error introduced during the merging of adjacent measurements depends on velocity and time offsets. Accepting this error might be beneficial to maintain a connected graph for joint estimation rather than preserving disconnected vertices. At high dynamics, graph candidates with the least MOM value tend to replicate the baseline graph structure, which is logical since the excessive clustering can degrade the map quality. 

2. The relative transformations obtained with the scan-matching algorithm may inaccurately estimate displacements, and the covariance of the corresponding measurement's noise can be assessed incorrectly or change dynamically. When merging clusters with these measurements, the aggregation procedure partially mitigates these errors through averaging.

Graph connectivity is a necessary but not sufficient condition for achieving globally consistent estimations. Fig.~\ref{fig:error-timeshift-problem} illustrates a challenge encountered in both Minimal Time Shift and Minimal Solver Error scenarios. Starting the motion from the stationary pose \(X_1\), the initial scans from both LiDARs can be combined. Two distinct pose chains (red and purple) are generated and connected by odometry factors. These factors do not influence the total time shift or solver error, as they connect scans at corresponding time instances and introduce independent equations for new variables. The GPS factors associated with poses \(X_2, X_6\) can compensate for the drift in odometry measurements of the LiDAR 1 chain but not in the chain of LiDAR 2. This problem diminishes promising compression rates due to high Root Mean Squared RPEs, computed using Eq.~\ref{eq:rpe} and presented in Table~\ref{tab:results}. In contrast, our method detects and resolves such inconsistencies by evaluating the resulting maps, ensuring globally consistent estimations and accurate map reconstruction.

%% file: chapters/conclusion.tex
\section{Conclusion}
\label{sec:conclusion}

This paper introduces a novel approach to factor graph-based SLAM that addresses the graph connectivity problem arising from aperiodic and asynchronous sensor measurements while ensuring the integration of all available data. The proposed method generates sub-optimal factor graphs for joint variable estimation and is applicable for fusing information from any sesnor. Experimental validation in an urban environment under realistic traffic dynamics demonstrates that our method compresses the factor graph by approximately \(\sim\)30\% on average without visually distinguishable loss of map quality. This is achieved by clustering adjacent measurements and selecting a sub-optimal graph topology based on a Mutually Orthogonal Metric. The evaluation criterion is not limited to the chosen metric and can be substituted with alternative schemes tailored to different applications.
Future work will focus on developing a learning-based strategy to identify optimal graph candidates, which could lower the computational complexity to polynomial levels depending on the neural network architecture. Another direction involves addressing the limitations of current evaluation criteria to bring the graph topology closer to optimality. The contributions of this work enable the generation of a dataset with labeled sub-optimal graph candidates, thereby facilitating the training of such a neural network.

%% file: chapters/appendix.tex
\section{Appendix}
\label{sec:appendix}

\subsection{Upper bound on the total number of combinations}

For a sequence $\mathcal{C}^N = \{C_1, C_2, \dots, C_N\}$ of \( N \) clusters and no continuous measurements \( \mathcal{M}^{I} \), there are no more than \( \mathcal{T} = \sum_{1}^{N}{\mathcal{D}_i} = 2^{N-1} \) possible merging combinations, where \( \mathcal{D} \) is a vector of binomial coefficients of the \( N\text{-}1 \) row of Pascal's triangle.

For a sequence \( \mathcal{C}^{N} \) and an arbitrary number of pre-integratable continuous measurements \( \mathcal{M}^{I} \) of the same type, located between core measurements \( \mathcal{M}^{C} \) and used to create non-intersecting connections, there are no more than \( 2^{N-2} \) different combinations with connections.

For N core measurements \( \mathcal{M}^{C} \)and an arbitrary number of pre-integratable continuous measurements there are in total:

\begin{equation}
    \label{eq:total_combinations}
    \mathcal{T} = \sum_{i=0}^{M} C(M, i) \cdot 2^{M-(i+1)},
\end{equation}
where \( M = N-1 \) and \( C(M, i) \) is the \( i \)-th binomial coefficient of the \( M \)-th row. By rewriting \( 2^{M-(i+1)} \) as \( 2^{M-(i+1)} = 2^{M} 2^{i+1} = \frac{1}{2} \cdot 2^{M} 2^{-i} \) and substituting back into Eq.~\ref{eq:total_combinations}:

\begin{equation}
    \begin{aligned}
        \mathcal{T} &= \sum_{i=0}^{M} C(M, i) \cdot 2^{M-(i+1)} = \sum_{i=0}^{M} C(M, i) \cdot 2^{M-1} \cdot 2^{-i} \\
        &= 2^{M-1} \sum_{i=0}^{M} C(M, i) \cdot \left( \frac{1}{2} \right)^{i}.
    \end{aligned}
\end{equation}

Applying the Binomial Theorem the total number of all possible combinations:

\begin{equation}
    \label{eq:result_T}
    \begin{aligned}
        \mathcal{T} &= 2^{M-1} \sum_{i=0}^{M} C(M, i) \cdot \left( \frac{1}{2} \right)^{i} = 2^{M-1} \left( 1 + \frac{1}{2} \right)^{M} \\
        &= 2^{M-1} \left( \frac{3}{2} \right)^{M} = \frac{3^M}{2} = \frac{3^{N-1}}{2}.
    \end{aligned}
\end{equation}

The total number of combinations is an integer number; however, Eq.~\ref{eq:result_T} produces a fractional result. This value should be rounded up to the nearest larger integer because a combination involving two clusters has one possible connection between them. A combination with a single cluster containing all core measurements has no connections but is still considered to be valid, produces factors that may connect the cluster with others. A critical time moment \( T \), referred to in the literature as a "coincidence peak" or "synchronization point," occurs when all sensors sample measurements at approximately the same time, which can be roughly estimated as the Least Common Multiple (LCM) of their periods. In this case, the number of unique measurements reaches its maximum. For a fixed time window \( \Delta{T} \) that includes the time moment \( T \) of a "coincidence peak," the number of measurements is finite and bounded by some constant \( K \) for a fixed number of sensors. Hence, the number of unique combinations calculated using Eq.~\ref{eq:result_T} is also finite and bounded by the constant \( 2^{N-1} \approx \text{const.} \) for \( \mathcal{M}^{C} \) and \( \frac{3^{N-1}}{2} \approx \text{const.} \) for both \( \mathcal{M}^{C} \) and \( \mathcal{M}^{I} \).

%% file: references.bib
@article {graphs_vs_kalman,
	author = {Taylor, Clark and Gross, Jason},
	title = {Factor Graphs for Navigation Applications: A Tutorial},
	volume = {71},
	number = {3},
	elocation-id = {navi.653},
	year = {2024},
	doi = {10.33012/navi.653},
	publisher = {Institute of Navigation},
	issn = {0028-1522},
	journal = {NAVIGATION: Journal of the Institute of Navigation}
}

@article{factor_graph,
	author = {Dellaert, Frank and Kaess, Michael},
	year = {2017},
	month = {01},
	pages = {1-139},
	title = {Factor Graphs for Robot Perception},
	volume = {6},
	journal = {Foundations and Trends in Robotics},
	doi = {10.1561/2300000043}
}

@inproceedings{b_spline_1,
	author={Furgale, Paul and Barfoot, Timothy D. and Sibley, Gabe},
	booktitle={2012 IEEE International Conference on Robotics and Automation}, 
	title={Continuous-time batch estimation using temporal basis functions}, 
	year={2012},
	volume={},
	number={},
	pages={2088-2095},
	doi={10.1109/ICRA.2012.6225005}}

@inproceedings{b_spline_2,
	author={Liu, Minjie and Huang, Shoudong and Dissanayake, Gamini and Kodagoda, Sarath},
	booktitle={2010 IEEE/RSJ International Conference on Intelligent Robots and Systems}, 
	title={Towards a consistent SLAM algorithm using B-Splines to represent environments}, 
	year={2010},
	volume={},
	number={},
	pages={2065-2070},
	doi={10.1109/IROS.2010.5649703}}

@article{b_spline_3,
	author={Park, Chanoh and Moghadam, Peyman and Williams, Jason L. and Kim, Soohwan and Sridharan, Sridha and Fookes, Clinton},
	journal={IEEE Transactions on Robotics}, 
	title={Elasticity Meets Continuous-Time: Map-Centric Dense 3D LiDAR SLAM}, 
	year={2022},
	volume={38},
	number={2},
	pages={978-997},
	doi={10.1109/TRO.2021.3096650}}

@article{b_spline_4,
	author={Lv, Jiajun and Lang, Xiaolei and Xu, Jinhong and Wang, Mengmeng and Liu, Yong and Zuo, Xingxing},
	journal={IEEE/ASME Transactions on Mechatronics}, 
	title={Continuous-Time Fixed-Lag Smoothing for LiDAR-Inertial-Camera SLAM}, 
	year={2023},
	volume={28},
	number={4},
	pages={2259-2270},
	doi={10.1109/TMECH.2023.3241398}}

@article{measurement_interpolation,
	title={Range-only slam with interpolated range data},
	author={Kehagias, Athanasios and Djugash, Joseph and Singh, Sanjiv},
	journal={Robotics Institute, Pittsburgh, PA, Tech. Rep. CMU-RI-TR-06-26},
	year={2006},
	publisher={Citeseer}
}

@inproceedings{ptam,
	author={Georg Klein, David Murray},
	booktitle={2007 6th IEEE and ACM International Symposium on Mixed and Augmented Reality}, 
	title={Parallel Tracking and Mapping for Small AR Workspaces}, 
	year={2007},
	pages={225-234},	
	doi={10.1109/ISMAR.2007.4538852}}

@inproceedings{bundle_adjustment,
	author    = {Bill Triggs and Philip F. McLauchlan and Richard I. Hartley and Andrew W. Fitzgibbon},
	title     = {Bundle Adjustment --- A Modern Synthesis},
	booktitle = {Vision Algorithms: Theory and Practice},
	year      = {2000},
	publisher = {Springer Berlin Heidelberg},
	address   = {Berlin, Heidelberg},
	pages     = {298--372},
	isbn      = {978-3-540-44480-0}
}

@inproceedings{kinect_fusion,
	author={Newcombe, Richard A. and Izadi, Shahram and Hilliges, Otmar and Molyneaux, David and Kim, David and Davison, Andrew J. and Kohi, Pushmeet and Shotton, Jamie and Hodges, Steve and Fitzgibbon, Andrew},
	booktitle={2011 10th IEEE International Symposium on Mixed and Augmented Reality}, 
	title={KinectFusion: Real-time dense surface mapping and tracking}, 
	year={2011},
	volume={},
	number={},
	pages={127-136},
	doi={10.1109/ISMAR.2011.6092378}}

@article{CFS,
	author = {Williams, S. and Indelman, V. and Kaess, M. and Roberts, R. and Leonard, J. J. and Dellaert, F.},
	title = {Concurrent Filtering and Smoothing: A Parallel Architecture for Real-Time Navigation and Full Smoothing},
	journal = {The International Journal of Robotics Research},
	volume = {33},
	year = {2014},
	month = {July},
	publisher = {Sage Publications},
	doi = {10.1177/0278364914543791},
	issn = {0278-3649}
}

@article{square_root_sam,
	author    = {Frank Dellaert},
	title     = {Square Root SAM: Simultaneous Localization and Mapping via Square Root Information Smoothing},
	journal   = {The International Journal of Robotics Research},
	volume    = {25},
	number    = {12},
	pages     = {1181--1203},
	year      = {2006},
	publisher = {SAGE Publications},
	doi       = {10.1177/0278364906072768}
}

@article{map_entropy,
	author    = {Henri Kretzschmar and Cyrill Stachniss},
	title     = {Information-theoretic compression of pose graphs for laser-based SLAM},
	journal   = {The International Journal of Robotics Research},
	volume    = {31},
	number    = {11},
	pages     = {1219--1230},
	year      = {2012},
	doi       = {10.1177/0278364912455072},
	publisher = {SAGE Publications}
}

@article{GNR,
	author={Carlevaris-Bianco, Nicholas and Kaess, Michael and Eustice, Ryan M.},
	journal={IEEE Transactions on Robotics}, 
	title={Generic Node Removal for Factor-Graph SLAM}, 
	year={2014},
	volume={30},
	number={6},
	pages={1371-1385},
	doi={10.1109/TRO.2014.2347571}}

@article{factor_decent_sparcification,
	author={Vallvé, Joan and Solà, Joan and Andrade-Cetto, Juan},
	journal={IEEE Robotics and Automation Letters}, 
	title={Graph SLAM Sparsification With Populated Topologies Using Factor Descent Optimization}, 
	year={2018},
	volume={3},
	number={2},
	pages={1322-1329},
	doi={10.1109/LRA.2018.2798283}}

@article{NFR,
	author    = {Marco Mazuran and Wolfram Burgard and Gian Diego Tipaldi},
	title     = {Nonlinear factor recovery for long-term SLAM},
	journal   = {The International Journal of Robotics Research},
	volume    = {35},
	number    = {1-3},
	pages     = {50--72},
	year      = {2016},
	doi       = {10.1177/0278364915581629},
	publisher = {SAGE Publications}
}

@article{se_sync,
  author    = {Rosen, D. M. and Carlone, L. and Bandeira, A. S. and Leonard, J. J.},
  title     = {SE-Sync: A certifiably correct algorithm for synchronization over the special Euclidean group},
  journal   = {The International Journal of Robotics Research},
  volume    = {38},
  number    = {2-3},
  pages     = {95--125},
  year      = {2019},
  doi       = {10.1177/0278364918784361}
}

@article{certifiable_slam,
  author={Holmes, Connor and Barfoot, Timothy D.},
  journal={IEEE Robotics and Automation Letters}, 
  title={An Efficient Global Optimality Certificate for Landmark-Based SLAM}, 
  year={2023},
  volume={8},
  number={3},
  pages={1539-1546},
  doi={10.1109/LRA.2023.3238173}}

@inproceedings{semantic_segm_slam,
  author={Wang, Yukun and Duan, Xiaojie and Sun, Yukuan and Wang, Jianming},
  booktitle={2021 4th International Conference on Pattern Recognition and Artificial Intelligence (PRAI)}, 
  title={A Visual SLAM Algorithm Based on Image Semantic Segmentation in Dynamic Environment}, 
  year={2021},
  volume={},
  number={},
  pages={401-405},
  doi={10.1109/PRAI53619.2021.9550800}}

@inproceedings{depth_slam,

  author={Chen, Yichen and Pan, Yuqi and Liu, Ruyu and Zhang, Haoyu and Zhang, Guodao and Sun, Bo and Zhang, Jianhua},
  booktitle={2024 27th International Conference on Computer Supported Cooperative Work in Design (CSCWD)}, 
  title={360ORB-SLAM: A Visual SLAM System for Panoramic Images with Depth Completion Network}, 
  year={2024},
  pages={717-722},
  doi={10.1109/CSCWD61410.2024.10580875}
  }

@ARTICLE {kaist,
    author = { Jinyong Jeong and Younggun Cho and Young-Sik Shin and Hyunchul Roh and Ayoung Kim },
    title = { Complex Urban Dataset with Multi-level Sensors from Highly Diverse Urban Environments },
    journal = { International Journal of Robotics Research },
    year = { 2019 },
    volume = { 38 },
    number = { 6 },
    pages = { 642--657 },
}

@article{mom,
  title={Be your own Benchmark: No-Reference Trajectory Metric on Registered Point Clouds},
  author={Anastasiia Kornilova and Gonzalo Ferrer},
  journal={2021 European Conference on Mobile Robots (ECMR)},
  year={2021},
  pages={1-8},
}

@article{kiss_icp,
  author    = {Vizzo, Ignacio and Guadagnino, Tiziano and Mersch, Benedikt and Wiesmann, Louis and Behley, Jens and Stachniss, Cyrill},
  title     = {{KISS-ICP: In Defense of Point-to-Point ICP -- Simple, Accurate, and Robust Registration If Done the Right Way}},
  journal   = {IEEE Robotics and Automation Letters (RA-L)},
  pages     = {1029--1036},
  doi       = {10.1109/LRA.2023.3236571},
  volume    = {8},
  number    = {2},
  year      = {2023},
  codeurl   = {https://github.com/PRBonn/kiss-icp}
}

@inproceedings{imu_preint,
	author    = {Forster, Christian and Carlone, Luca and Dellaert, Frank and Scaramuzza, Davide},
	booktitle = {Robotics: Science and Systems},
	title     = {IMU Preintegration on Manifold for Efficient Visual-Inertial Maximum-a-Posteriori Estimation},
	year      = {2015},
	ee        = {https://doi.org/10.15607/RSS.2015.XI.006},
	isbn      = {978-0-9923747-1-6}
}

@misc{gtsam,
  author       = {Frank Dellaert and GTSAM Contributors},
  title        = {borglab/gtsam},
  month        = May,
  year         = 2022,
  publisher    = {Georgia Tech Borg Lab},
  version      = {4.2a8},
  doi          = {10.5281/zenodo.5794541},
  url          = {https://github.com/borglab/gtsam}
}

@inproceedings{moduslam,
  author={Griguletskii, Mark and Osinenko, Pavel},
  booktitle={2025 IEEE 26th International Conference of Young Professionals in Electron Devices and Materials (EDM)}, 
  title={Moduslam: A Modular Framework for Factor Graph-Based Localization and Mapping}, 
  year={2025},
  pages={1210-1215},
  doi={10.1109/EDM65517.2025.11096772}
  }
